\documentclass[10pt,twocolumn,letterpaper]{article}

\usepackage[pagenumbers]{cvpr} 
\usepackage{graphicx}
\usepackage{amsmath}
\usepackage{amssymb}
\usepackage{booktabs}
\usepackage[percent]{overpic}
\usepackage{color}
\usepackage{colortbl}  
\usepackage{xcolor}
\usepackage{array}
\usepackage{bbding}
\definecolor{hollywoodcerise}{rgb}{0.96, 0.0, 0.63}
\definecolor{lasallegreen}{rgb}{0.03, 0.47, 0.19}
\definecolor{hanpurple}{rgb}{0.32, 0.09, 0.98}
\definecolor{green(pigment)}{rgb}{0.0, 0.65, 0.31}
\usepackage{multirow}
\newcommand{\aka}{\textit{a.k.a.}}

\definecolor{cvprblue}{rgb}{0.21,0.49,0.74}
\usepackage[pagebackref,breaklinks,colorlinks,citecolor=cvprblue]{hyperref}

\def\paperID{} 
\def\confName{CVPR}
\def\confYear{2024}

\title{Image Anything: Towards Reasoning-coherent and Training-free Multi-modal Image Generation}

\author{Yuanhuiyi Lyu$^{1}$ \quad Xu Zheng$^{1}$ \quad Lin Wang$^{1}$$^{,2}$\thanks{Corresponding author.}\\
$^{1}$AI Thrust, HKUST(GZ) \quad $^{2}$Dept. of CSE, HKUST 
\\
{\tt\small yuanhuiyilv@hkust-gz.edu.cn, xzheng287@hkust-gz.edu.cn, linwang@ust.hk}
\\
\tt \small Project Page: \url{https://vlislab22.github.io/ImageAnything/}
}

\begin{document}

\twocolumn[{%
\renewcommand\twocolumn[1][]{#1}%
\maketitle
\begin{center}
\centering
\vspace{-16pt}
\begin{overpic}[width=\textwidth]{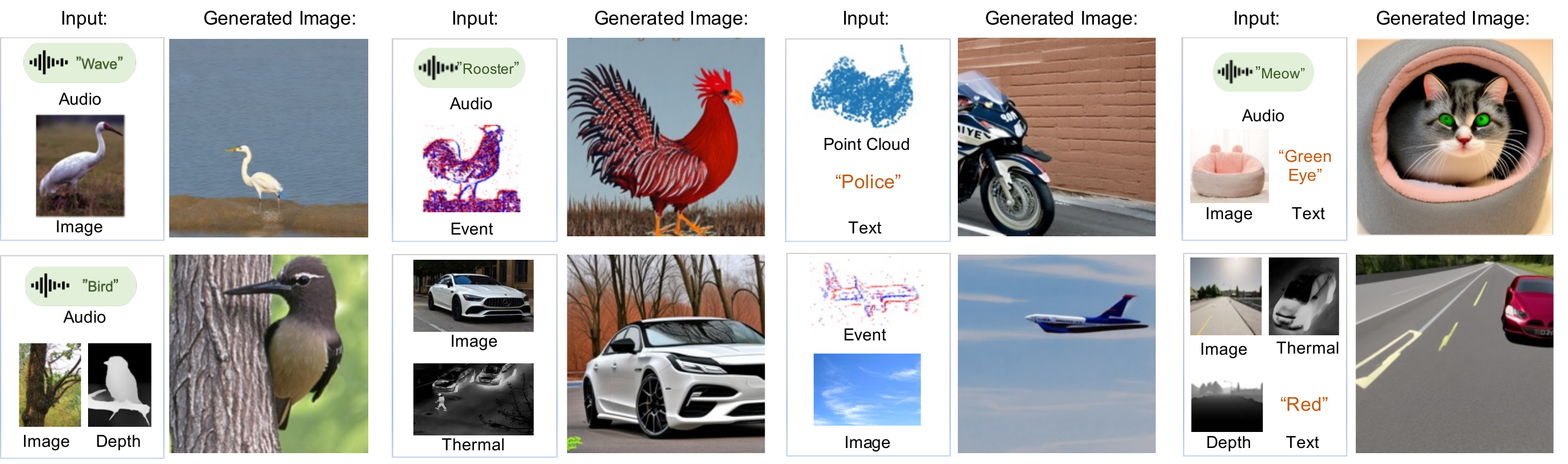}
\end{overpic}
\captionof{figure}{
 We introduce \textbf{ImgAny}, a novel training-free multi-modal image generation framework that can mimic human reasoning and generate high-quality images from any modalities, ranging from language, audio, to vision modalities, such as image, point cloud, thermal, depth, and event data.
  Our ImgAny subtly integrates and harmonizes the modalities at the entity and attribute levels without specific tuning across them to generate visually appealing images.
 }
\label{fig:teaser}
\end{center}%
}]
\renewcommand{\thefootnote}{} 
\footnotetext{$^\dagger$Corresponding author.}
\begin{abstract}
The multifaceted nature of human perception and comprehension indicates that, when we think, our body can naturally take any combination of senses, \aka, modalities, and form a beautiful picture in our brain. For example, when we see a cattery and simultaneously perceive the cat's purring sound, our brain can construct a picture of a cat in the cattery. Intuitively, generative AI models should hold the versatility of humans and be capable of generating images from \textbf{any} combination of modalities efficiently and collaboratively.  
This paper presents \textbf{ImgAny}, a novel end-to-end multi-modal generative model that 
can mimic human reasoning and generate high-quality images. Our method serves as the \textbf{first} attempt in its capacity of \textbf{efficiently} and \textbf{flexibly} taking any combination of \textbf{seven} modalities, ranging from language, audio to vision modalities, including image, point cloud, thermal, depth, and event data. 
Our key idea is inspired by human-level cognitive processes and involves the integration and harmonization of multiple input modalities at both the entity and attribute levels \textit{without specific tuning across modalities}. Accordingly, our method brings two novel \textbf{training-free} technical branches:
\textbf{1)} \textbf{Entity Fusion Branch} ensures the coherence between inputs and outputs. It extracts entity features from the multi-modal representations powered by our specially constructed entity knowledge graph;
\textbf{2) Attribute Fusion Branch} adeptly preserves and processes the attributes. It efficiently amalgamates distinct attributes from diverse input modalities via our proposed attribute knowledge graph.
Lastly, the entity and attribute features are adaptively fused as the conditional inputs to the pre-trained Stable Diffusion model for image generation.
Extensive experiments under diverse modality combinations demonstrate its exceptional capability for visual content creation. 

\end{abstract}

\section{Introduction}
Multi-modal image generation has recently received great attention for content creation in the domains of media, art, Metaverse, \etc. Aiming at mimicking human imagination, it elucidates the importance of enabling generative models to correlate the multi-modal attributes, \eg, visual, text, and audio descriptions, to generate images~\cite{wu2023next,openai2023gpt4,touvron2023llama}.
Early methods primarily facilitated image generation from singular modal inputs, such as image~\cite{lu2018image,choi2018stargan} or text~\cite{li2019controllable,qiao2019mirrorgan} or audio~\cite{yang2020diverse,zelaszczyk2022audio}. 
In pursuit of enhanced control over generation, a prevalent direction has been focused on image generation under the cross-modal guidance by integrating two modalities, such as image and text~\cite{zhang2023adding,brooks2023instructpix2pix,huang2023region,ham2023modulating} or image and audio~\cite{sanguineti2022unsupervised,sung2023sound}.
Nonetheless, these approaches encounter constraints in real-world contexts defined by more complex modality inputs. 
As a result, there is a growing need for versatile models that can effectively process a wide range of input modalities, offering a more accurate representation of human perception and imagination.

Recently, it has witnessed dramatic advancement thanks to the development of diffusion models~\cite{rombach2022high} and neural rending techniques~\cite{mildenhall2021nerf,muller2022instant,wang2021neus, jiang2023sdf}. A representative effort of recent literature is CoDi~\cite{tang2023any}, which can process various combinations of modalities, such as image, text, and audio to generate images.
However, these methods fail to generate content with natural sensing integration from multi-modal inputs like humans image a scene.
For instance, as shown in Fig.~\ref{fig:teaser}, when we observe a cattery, read text describing green eyes, and simultaneously hear a cat's purring sound, a cohesive mental image of a cat with green eyes situated within the cattery is naturally formed in our mind. Intuitively, \textit{the generative models should hold the versatility of humans and generate high-quality images from any combination of modalities efficiently and collaboratively}. Another limitation is that RGB image is not the only visual representation of the world, and other vision modalities, such as thermal, depth, event, and point cloud, should be incorporated into the multi-modal generative models. Fig.~\ref{fig:teaser} demonstrates that combining text, audio, and other modalities, \eg, thermal, can also generate high-quality images.  

\begin{figure*}[ht!]
    \centering
    \includegraphics[width=\textwidth]{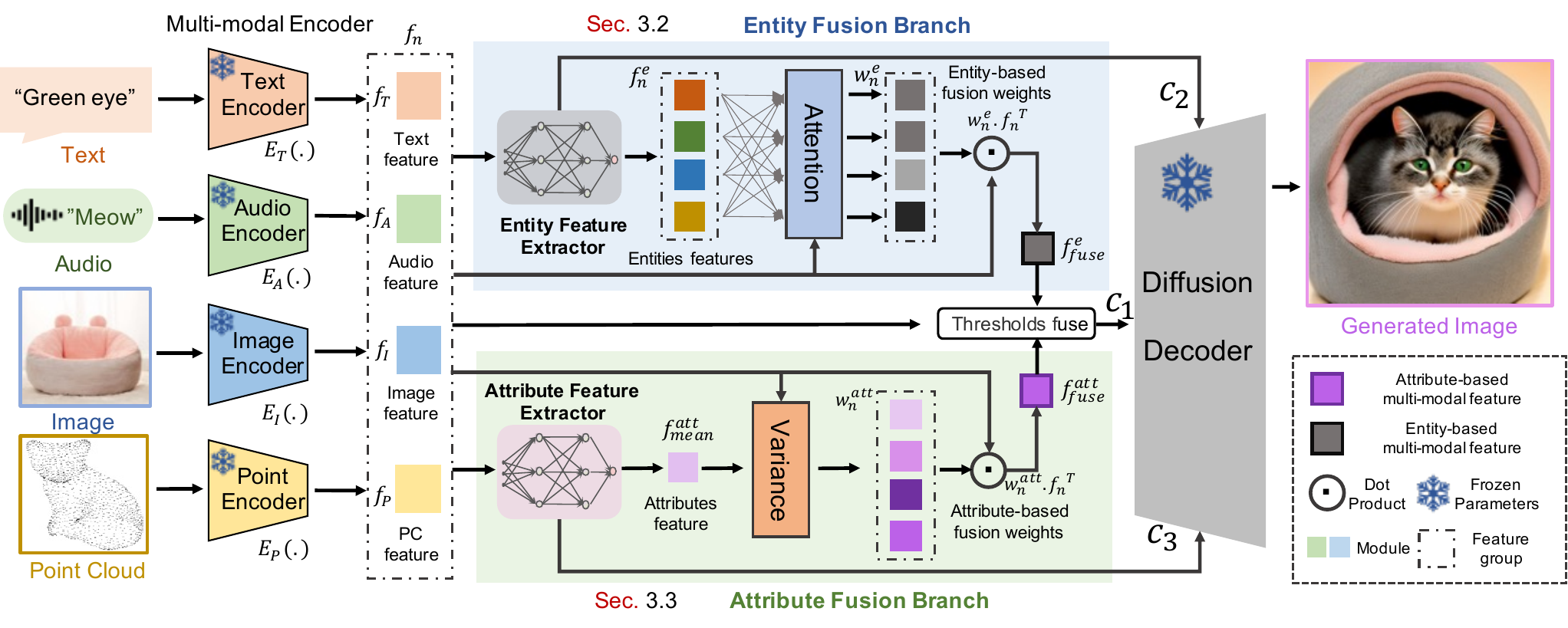}
    \caption{
    The overall framework of ImgAny includes: \textbf{1)} Multi-modal Encoder, \textbf{2)} Entity Fusion Branch, and \textbf{3)} Attribute Fusion Branch. 
    }
    \label{fig:overall}
\end{figure*}

In this paper, we present \textbf{ImgAny}, a novel end-to-end multi-modal generative model that can mimic human perception and imagination and generate high-quality images for content creation efficiently and flexibly, as depicted in Fig.~\ref{fig:overall}. ImgAny serves as the \textbf{first} attempt to take \textbf{any} combination of \textbf{seven} modalities, ranging from language, audio, to vision modalities, such as image, point cloud, thermal, depth, and event data.
\textit{Our key idea is inspired by human-level cognitive processes and involves the integration and harmonization of multiple input modalities at both the entity and attribute levels without specific tuning across modalities}. 
This is achieved by feature extraction with the multi-modal vision-language models~\cite{guo2023point} and our proposed two novel training-free technical branches, \textit{i.e.}, the entity fusion branch and the attribute fusion branch.

Specifically, the \textbf{entity fusion branch} is proposed to maintain coherence between inputs and outputs by integrating an external entity knowledge graph. 
This entity knowledge graph is built on the entity nouns in WordNet~\cite{miller1995wordnet} to offer insights into entity interrelationships, enhancing contextual alignment in generated images and ensuring accurate depiction of entities' spatial and relational aspects.
Meanwhile, the \textbf{attribute fusion branch} is designed to merge diverse attribute features drawn from all input modalities through an attribute knowledge graph.
This knowledge graph is underpinned by the comprehensive collection of attribute adjectives from WordNet~\cite{miller1995wordnet} and plays a crucial role in broadening the range of attributes represented in the generated images. The integration of the attribute knowledge graph facilitates a more nuanced and varied portrayal of attributes, closely aligning with the attribute details present in the multi-modal data.
Lastly, a frozen Stable Diffusion~\cite{rombach2022high} is utilized to generate images with the guidance of the fused multi-modal feature conditions.

We conducted extensive experiments to validate the efficacy and superiority of ImgAny. 
Quantitatively, the metrics underscore ImgAny's advanced capabilities in accurately interpreting and integrating diverse modal inputs to produce high-quality images (See Tab.~\ref{fig: com_2}). Qualitatively, the generated images exhibit remarkable fidelity, reflecting ImgAny's proficiency in emulating human-level reasoning and creativity. 
Both quantitative and qualitative results show that our ImgAny outperforms the prior methods in image generation from arbitrary combinations of modalities. 
Also, we perform human evaluation for ImgAny, and the results reveal that ImgAny can generate reasoning-coherent and high-quality images (See Fig.~\ref{fig: human}). 

In summary, our main contributions can be summarized as follows: 
(\textbf{I}) We propose ImgAny, the \textit{first} training-free end-to-end multi-modal generative model, capable of taking arbitrary combinations of \textit{seven} modalities; 
(\textbf{II}) We introduce two novel technical branches to integrate and harmonize multi-modal inputs at both entity and attribute levels \textit{without specific tuning} across modalities; (\textbf{III}) Both \textit{numerical and human-level} evaluations demonstrate that our ImgAny can generate reasoning-coherent and high-quality images from any combination of modalities.

\section{Related Work}
\noindent\textbf{Multi-modal Image Generation.}
Early image generation methods primarily focused on generating images from single-modal inputs, such as images~\cite{lu2018image,choi2018stargan}, text~\cite{li2019controllable,qiao2019mirrorgan}, or audio~\cite{yang2020diverse,zelaszczyk2022audio}. However, human perception encompasses diverse senses that process information from language, vision, and sound, collaborating to comprehensively understand the environment. To enhance control over generation and mimic human imagination, efforts have focused on cross-modal guidance in image generation by integrating two modalities~\cite{sanguineti2022unsupervised,sung2023sound,zhang2023adding,brooks2023instructpix2pix,huang2023region,ham2023modulating}.
Despite these advancements, dual-modality inputs fall short in real-world scenarios with more intricate inputs. Consequently, LLMs, driven solely by language, have been enhanced with additional capabilities across visual, audio, and other modalities~\cite{li2023blip,zhu2023minigpt,zhang2023video,su2023pandagpt,zhang2023speechgpt,wu2023next, hong20233d, jiang2023motiongpt, maaz2023video, zhang2023llama} for multi-modal image generation. Recent initiatives~\cite{wu2023visual,shen2023hugginggpt} aim to integrate LLMs, developing systems capable of achieving 'any-to-any' multi-modal generation. However, their reliance on LLM-generated texts as a communication medium between different modules introduces potential inaccuracies.
Another research direction focuses on end-to-end learning. A closely related effort in our problem setting is CoDi~\cite{tang2023any}, capable of handling any combination of input modalities, including language, image, video, and audio. However, CoDi~\cite{tang2023any} achieves the fusion of multi-modal features with user-defined hand-crafted weights, allowing only parallel cross-modal feeding and generation. This approach is less likely to achieve human-like deep reasoning of multi-modal input, as illustrated in Fig.~\ref{fig: com_1}.
Recent human-like multi-modal systems, such as NExT-GPT~\cite{wu2023next}, have been developed to achieve semantic understanding, reasoning, and generation using computationally-demanding LLMs. However, these systems are tailored for online dialogue and require text inputs exclusively.
In contrast, our ImgAny serves as the first attempt to take any combination of seven modalities, ranging from language and audio to five vision modalities. It achieves reasoning-coherent and training-free image generation through the lens of entity and attribute analysis with significantly reduced model parameters and inference costs (see Tab.~\ref{tab: campare_1}), paving the way for a more nuanced, human-like reasoning and imagination capability.

\noindent\textbf{Multi-modal Vision-Language Models.}
Recent research endeavors have undertaken a comprehensive exploration of multi-modal vision-language models, which are based on the unified vision-language representation space, built by the CLIP style large vision-language models~\cite{radford2021learning, li2022blip, wei2023iclip, liu2022universal, luo2022clip4clip, xue2022clip, zheng2023deep}.
These methods adapt one~\cite{zhou2023clip,zhang2022pointclip, zhu2022pointclip,guzhov2022audioclip,mahmud2023ave,fang2021clip2video} or more~\cite{girdhar2023imagebind, guo2023point} modalities to the image representation space to align multiple visual modalities. ImageBind~\cite{girdhar2023imagebind} is a representative work that expands multi-modal alignment by leveraging the binding property of images to learn a single shared representation space. Based on ImageBind, PointBind~\cite{guo2023point} aligns the point cloud to the image representation space with cross-modal correlation alignment. In our work, we utilize the PointBind as our multi-modal encoder backbone model to obtain the multi-modal features.

\section{The Proposed ImgAny}
\subsection{Overview}
The overall framework of ImgAny is depicted in Fig.~\ref{fig:overall}. 
Given $n$ modalities, our ImgAny includes $n$ encoders to extract multi-modal features. Then ImgAny fuses the obtained features with the entity fusion branch and attribute fusion branch at both entity and attribute levels, respectively. 
Lastly, ImgAny utilizes the fused multi-modal feature ($C_1$), together with the knowledgeable texts ($C_2$ and $C_3$) extracted by entity feature extractor and attribute feature extractor as the input conditions of the Stable Diffusion~\cite{rombach2022high}. 
Overall, ImgAny has three key modules:\textbf{1)} multi-modal encoders,
for the multi-modal data \{Image, ..., Audio\}: $\{I, ..., A\}$, we utilize frozen encoders from PointBind~\cite{guo2023point} to extract features:
\begin{equation}
    f_{I}, ..., f_{A} = E_{I}(I), ..., E_{A}(A),
\end{equation}
where the $f_{I}, ..., f_{A}$ are the multi-modal features and $E_{I}(\cdot), ..., E_{A}(\cdot)$ are the encoders. Then we fuse the multi-modal features with the two following fusion branches;
\textbf{2)} Entity Fusion Branch (Sec.~\ref{sec:EFB}),
and \textbf{3)} Attribute Fusion Branch (Sec.~\ref{sec:AFB}). 



\begin{figure}[t!]
    \centering
    \includegraphics[width=\linewidth]{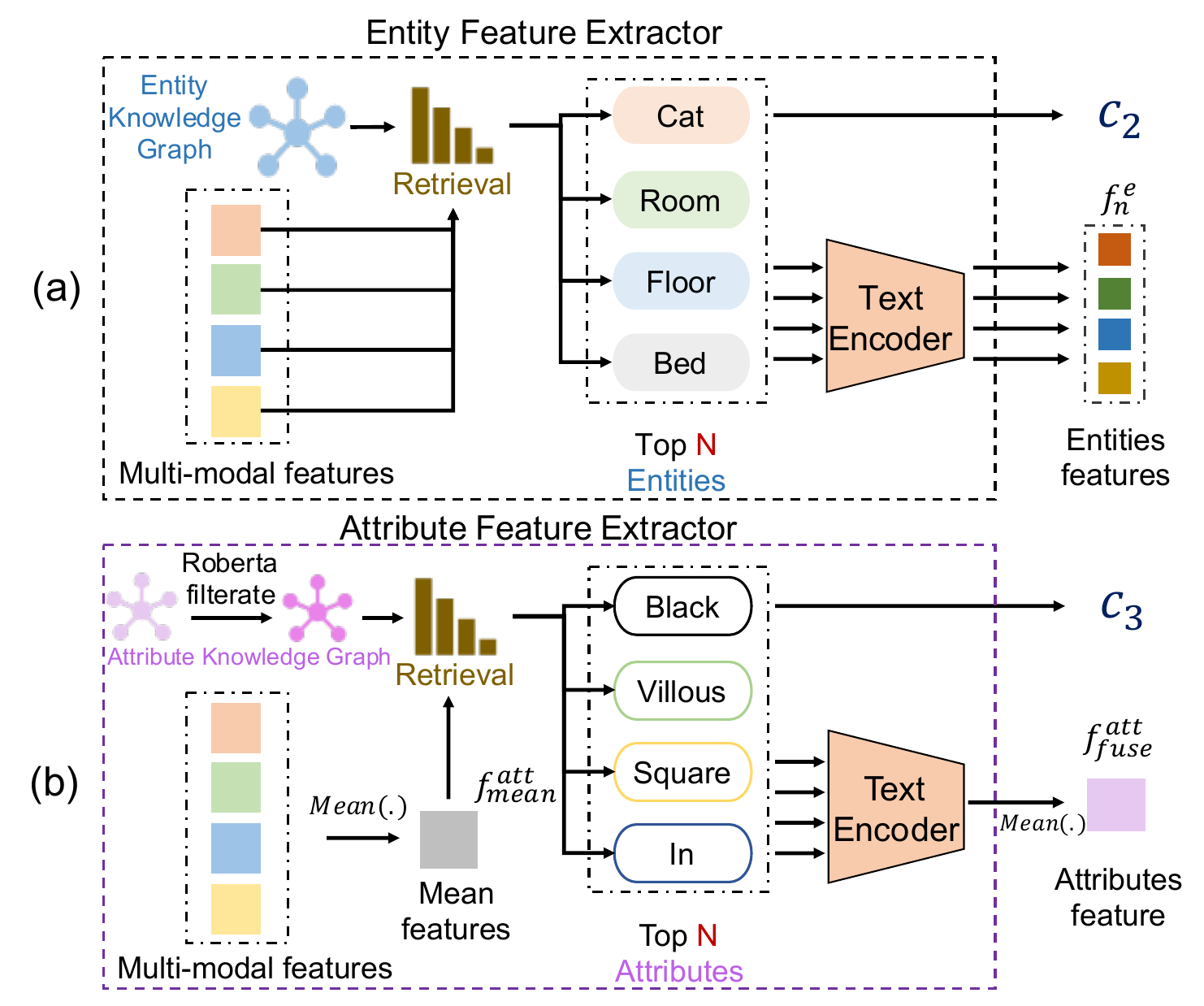}
    \caption{
    (a) Entity Feature Extractor;
    (b) Attribute Feature Extractor.
    }
    \label{fig:branch}
\end{figure}

\subsection{Entity Fusion Branch}
\label{sec:EFB} 
The entity fusion branch is first proposed to maintain coherence between inputs and outputs. To this end, an entity feature extractor is introduced to capture the entity features, as well as the generation condition $C_2$ for Stable Diffusion.
As shown in Fig.~\ref{fig:branch} (a), we first construct an entity knowledge graph by using all the entity nouns from WordNet~\cite{miller1995wordnet}. 
Subsequently, employing the text encoder from WordNet~\cite{guo2023point}, we extract the features of these entity nouns based on the specific text format {\fontfamily{qcr}\selectfont [a photo of a [nouns]]}. 
Upon extracting the features of these entity nouns, we determine the most relevant entity words correlating with the multi-modal features $f_{I}, ..., f_{A}$ by computing and ranking their cosine similarity with the nouns' features. This process is represented by the "Retrieval" in Fig.~\ref{fig:branch} (a). The nouns whose features exhibit the highest similarity to each multi-modal feature are chosen as the most pertinent entity words.

Once the most pertinent entity words are retrieved, exemplified by {\fontfamily{qcr}\selectfont [Cat]}, {\fontfamily{qcr}\selectfont [Room]}, {\fontfamily{qcr}\selectfont [Floor]}, and {\fontfamily{qcr}\selectfont [Bed]} in Fig.~\ref{fig:branch} (a), these words, encapsulating entity information from all input modalities, are concatenated to form condition $C_2$. 
We then proceed to extract the entity features with the following equation:
\begin{equation}
    f^e_{I}, ..., f^e_{A} = E_{T}(En_I), ..., E_{T}(En_A),
\end{equation}
where the $f^e_{I}, ..., f^e_{A}$ represent the extracted entity features for each modality, $En_I, ..., En_A$ represent the most related entity words and $E_{T}(\cdot)$ is the text encoder.
Subsequently, we infer the entity-based fusion weights with the multi-modal features by:
\begin{equation}
    w^e_{I}, ..., w^e_{A} = (\sum_{f^e=f^e_{I}}^{f^e_{A}} f_{I} \cdot f^e) / n, ..., (\sum_{f^e=f^e_{I}}^{f^e_{A}} f_{A} \cdot f^e) / n,
\end{equation}
where the $w^e_{I}, ..., w^e_{A}$ are the entity-based fusion weights and $n$ is the number of the most related entity words.
Lastly, powered by the weights, we fuse the multi-modal features and obtain the entity-based multi-modal feature $f_{fuse}^e$:
\begin{equation}
    f_{fuse}^e = (w^e_{I}, ..., w^e_{A}) \cdot (f_{I}, ..., f_{A})^T,
\end{equation}

\subsection{Attribute Fusion Branch}
\label{sec:AFB}
Despite ensuring entity information, it is also crucial to merge diverse attribute features drawn from all inputs.
Akin to the entity fusion branch, we propose an attribute feature extractor to derive attribute features and formulate the generation condition $C_3$ for Stable Diffusion. As depicted in Fig.~\ref{fig:branch} (b), we first construct an attribute knowledge graph by harnessing attribute adjectives from WordNet~\cite{miller1995wordnet}, while employing the pre-trained language model, Roberta~\cite{liu2019roberta}, to exclude abstract adjectives like 'great' and 'good'. 
We approach this exclusion as a binary classification task, utilizing the {\fontfamily{qcr}\selectfont [CLS]} token embedding for decision-making. After obtaining the attribute knowledge graph, we then utilize the text encoder to extract attribute adjectives's features.

Similar to the entity fusion branch, the mean of multi-modal features is used to retrieve the most pertinent attribute words among the attribute adjectives. 
The retrieval is also achieved by cosine similarity computing and ranking between the mean features $f_{mean}^{att}$ and the extracted attribute adjectives features. The adjectives whose features have larger similarities to the mean features are selected as the relevant attribute words.
These relevant attribute words, such as {\fontfamily{qcr}\selectfont [Balck]}, {\fontfamily{qcr}\selectfont [Villous]}, {\fontfamily{qcr}\selectfont [Square]}, and {\fontfamily{qcr}\selectfont [In]} in Fig.~\ref{fig:branch} (b), are then amalgamated to form condition $C_3$, following which we extract the attribute features by:
\begin{equation}
    f^{att}_{I}, ..., f^{att}_{A} = E_{T}(Att_I), ..., E_{T}(Att_A),
\end{equation}
where the $f^att_{I}, ..., f^att_{A}$ are the extracted attribute features for each modality and $Att_I, ..., Att_A$ are the retrieved most related attribute words.
Then, we calculate the attribute-based fusion weights via multi-modal features and the attribute features:
\begin{scriptsize}
\begin{equation}
    w^{att}_{I}, ..., w^{att}_{A} = (f_I - \frac{1}{n} \sum_{f^{att}=f^{att}_{I}}^{f^{att}_{A}}f^{att})^2, ..., (f_A - \frac{1}{n} \sum_{f^{att}=f^{att}_{I}}^{f^{att}_{A}}f^{att})^2 ,
\end{equation}
\end{scriptsize}
where the $w^{att}_{I}, ..., w^{att}_{A}$ are the entity-based fusion weights and $n$ is the number of the most related attribute words.
We then fuse the multi-modal features and obtain the attribute-based feature $f_{fuse}^{att}$:
\begin{equation}
    f_{fuse}^{att} = (w^{att}_{I}, ..., w^{att}_{A}) \cdot (f_{I}, ..., f_{A})^T.
\end{equation}
Lastly, we fuse entity-based multi-modal feature $f_{fuse}^{e}$ and attribute-based multi-modal feature $f_{fuse}^{att}$ to get the final feature $f_{fuse}$ and take it as the condition $C_1$ for Stable Diffusion decoder:
\begin{equation}
    f_{fuse} = T_{fuse}(f_{fuse}^{e}, f_{fuse}^{att}),
\end{equation}
where the $T_{fuse}$ is a threshold function in which $f_{fuse}^{e}$ and $f_{fuse}^{att}$ are added together with weights of 0 to 1 when the variance of the input distribution is less than a certain threshold 0.8, when it is greater than a certain threshold, the weight of $f_{fuse}^{e}$ is adjusted upward as 0.6 to ensure the stability of the entity information in the fused multi-modal feature.
Overall, ImgAny generates images with the obtained feature conditions $C_1$, $C_2$, and $C_3$ via the Stable Diffusion decoder, with the parameters of Stable Diffusion frozen.

\subsection{Implementation}

\noindent \textbf{Backbone Encoder:} 
ImgAny employs PointBind's~\cite{guo2023point} multi-modal encoders for feature extraction, while PointBind's text encoder is utilized for retrieving related word embeddings from knowledge graphs. 
\noindent \textbf{Generation Model:} We use frozen pre-trained Stable Diffusion V2.0~\cite{rombach2022high} as the generation decoder, which iteratively denoises from an initial Gaussian noise image with multi-modal feature conditions $\{C_1, C_2, C_3\}$.

\begin{figure*}[h!]
    \centering
    \includegraphics[width=\textwidth]{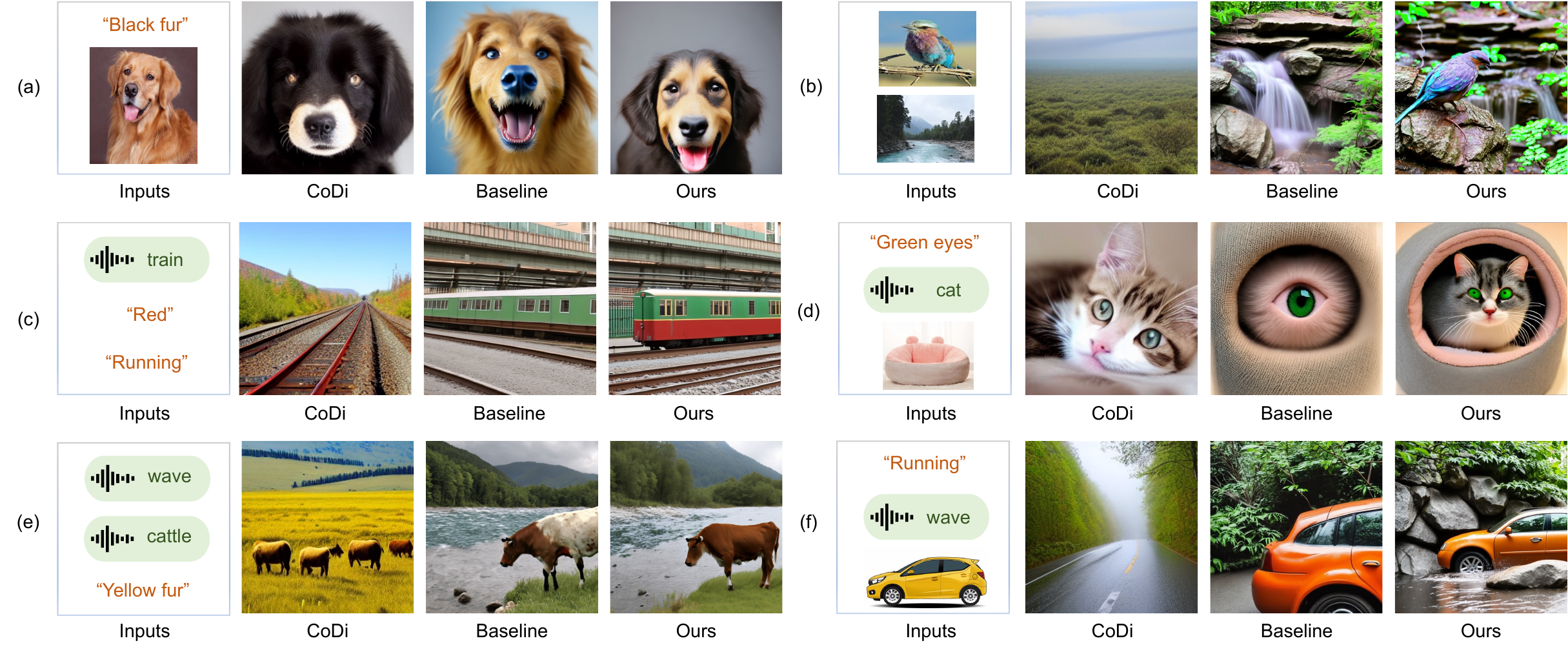}
    \caption{Qualitative comparison results. Compared with CoDi and baseline, our ImgAny shows better consistency and accuracy.}
    \label{fig: com_1}
\end{figure*}
\section{Experiment}
\subsection{Experimental Setup}
\noindent \textbf{Comparison Methods.} 
\textbf{1) CoDi}~\cite{tang2023any} is a generative model that is capable of generating images from texts, images, audios, and videos. Since the inputs of CoDi require the user to give weights for each modality, we defaulted to the same weights for each modality in the experiments. \textbf{2) baseline} model contacts PointBind~\cite{guo2023point} and Stable Diffusion and utilizes the mean of multi-modal features as the condition input for Stable Diffusion to generate images.

\noindent \textbf{Experiment Settings.} 
\textit{1) Image Generation from arbitrary Combinations of \{Text, Audio, Image\}:} We explore the generation capabilitieswith text, audio, and image as input modalities. 
\textit{2) Image Generation from arbitrary Combinations of \{Text, Audio, Image, Point Cloud, Thermal, Event, Depth\}:} 
Given that CoDi does not support these input modalities for image generation, our comparison primarily revolves around the baseline and our ImgAny.

\noindent \textbf{Datasets.} We utilize mainstream public datasets for quantitative evaluation including Flickr-30K~\cite{young2014image} for text-to-image generation, ESC-50~\cite{piczak2015esc} for audio-to-image generation, and FLIR V1~\cite{teledyne2018free} for thermal-to-image generation.

\noindent \textbf{Evaluation Metrics.}
For quantitative evaluation, we utilize two well-recognized metrics:
\textbf{1) FID Score}: Measures the similarity between feature distributions of generated and real images. For applications like Audio-to-Image and Thermal-to-Image generation, ImageBind~\cite{han2023imagebind} is used for feature extraction. 
\textbf{2) CLIP Score}: Assesses similarity between input and output features in the CLIP space, with a higher score indicating stronger semantic alignment between inputs and the generated image.

\subsection{Image Generation from Text, Audio, Image}
\noindent \textbf{Qualitative Comparison.} 
As shown in Fig.\ref{fig: com_1}, ImgAny surpasses CoDi\cite{tang2023any} and baseline in recognizing and interpreting diverse semantic content from input modalities. 
ImgAny adeptly preserves essential objects and attributes, accurately replicating entity features (\eg, the dog's form) and attribute features (\eg, fur color) from the input multi-modal conditions in the generated images. 
For example, in Fig.\ref{fig: com_1} (a), CoDi~\cite{tang2023any} fails to retain the specific morphology of the "dog," resulting in a generic representation of a "black fur dog," while the baseline maintains the shape and variety but lacks the capability to modify fur color. In contrast, ImgAny not only preserves the entity features but also accurately replicates attribute features like the fur color.

This superiority becomes more evident with increasing modality complexity, as depicted in Fig.~\ref{fig: com_1} (d). ImgAny is the only approach that successfully integrates all three input modalities - "green eyes," "cat," and "pink cat’s nest" - into a single coherent image. ImgAny also can integrate different entities from the same modality, as shown in Fig.~\ref{fig: com_1} (e), comparing with CoDi and baseline, ImgAny balances the entity "Cattle" and "river" supposed by the audio modality and replicates the attribute features "Yellow fur" accurately. Moreover, our ImgAny exhibits a human-like cognitive ability to logically associate the green eyes with the cat and contextualize the cat within the pink cattery, showcasing its advanced reasoning capabilities in image generation from any combination of various modalities.

\begin{table}[t!]
\renewcommand{\tabcolsep}{2pt}
\resizebox{\linewidth}{!}{
\begin{tabular}{lccccccc}
\toprule
\multicolumn{1}{c|}{Model}& \multicolumn{1}{c|}{\#Para} & \multicolumn{2}{c|}{T$\rightarrow$I} & \multicolumn{2}{c|}{A$\rightarrow$I} & \multicolumn{2}{c|}{Th$\rightarrow$I} \\ \cmidrule{3-8}
& \multicolumn{1}{|c}{}&\multicolumn{1}{|c|}{CLIP} & \multicolumn{1}{c|}{FID} & \multicolumn{1}{c|}{CLIP} & \multicolumn{1}{c|}{FID} & \multicolumn{1}{c|}{CLIP} & \multicolumn{1}{c|}{FID} \\ \midrule
CoDi~\cite{tang2023any} & 5.71 B & 32.68 & 170.77 & \textbf{25.24} & 320.21 & \tiny{\XSolidBrush} & \tiny{\XSolidBrush} \\
Baseline & 2.49 B & 33.19 & 202.02 & 20.06 & 250.49 & 13.08 & 342.95 \\
ImgAny & 2.49 B & \textbf{37.74} & \textbf{157.26} & 23.99 & \textbf{241.18} & \textbf{15.66} & \textbf{314.24}\\
\rowcolor{gray!10}$\Delta $ & -3.22 B & +4.55 & +13.51 & -1.25 & +9.31 & +2.58 & +28.71  \\
\bottomrule
\end{tabular}}
\caption{Quantitative comparison on the text-to-image (T$\rightarrow$I), audio-to-image (A$\rightarrow$I), and thermal-to-image (Th$\rightarrow$I) tasks. 
}
\vspace{-8pt}
\label{tab: campare_1}
\end{table}

\noindent \textbf{Quantitative Comparison.} 
As shown in Tab. \ref{tab: campare_1}, ImgAny shows better performance in generating images from arbitrary combinations of text, audio, and image modalities, achieving a reduction of 56 \% in the number of parameters compared to CoDi~\cite{tang2023any}. The ImgAny achieves a 1.96 \% improvement in the average CLIP score and a significant enhancement of 17.18 \% in the FID score. Furthermore, as illustrated in Tab. \ref{tab: campare_2}, we detail the CLIP scores for the generation cases depicted in Fig. \ref{fig: com_1}, highlighting the markedly stronger resemblance between images generated by ImgAny and the input multi-modal conditions, in comparison to those produced by CoDi~\cite{tang2023any} and baseline.
\begin{table}[t!]
\renewcommand{\tabcolsep}{3pt}
\resizebox{\linewidth}{!}{
\begin{tabular}{lccccccc}
\toprule
\multicolumn{1}{c|}{Model} & \multicolumn{1}{c|}{\#Param} & \multicolumn{1}{c|}{(a)} & \multicolumn{1}{c|}{(b)} & \multicolumn{1}{c|}{(c)} & \multicolumn{1}{c|}{(d)} & \multicolumn{1}{c|}{(e)} & \multicolumn{1}{c|}{(f)} \\ \midrule
CoDi~\cite{tang2023any} & 5.71 B &70.79  & 27.27 & 31.21 &  31.22 & 33.78 & 28.77\\
Baseline & 2.49 B & 86.28 & 41.83 & 48.96 &  25.74 & 49.19 & 43.69 \\
ImgAny & 2.49 B & \textbf{90.70} & \textbf{51.64} & \textbf{49.95} & \textbf{48.27} & \textbf{49.69} &\textbf{ 45.42} \\
\rowcolor{gray!10}$\Delta $ & -3.22 B & +4.42 & +9.81 & +0.99 &  +17.05 & +0.50 & +1.73 \\
\bottomrule
\end{tabular}}
\caption{The CLIP score of the cases shown in Fig.~\ref{fig: com_1} }
\label{tab: campare_2}
\end{table}

\begin{table}[t!]
\renewcommand{\tabcolsep}{6pt}
\resizebox{\linewidth}{!}{
\begin{tabular}{lcccccc}
\toprule
\multicolumn{1}{c|}{Model} & \multicolumn{1}{c|}{(a)} & \multicolumn{1}{c|}{(b)}& \multicolumn{1}{c|}{(c)}& \multicolumn{1}{c|}{(d)}& \multicolumn{1}{c|}{(e)}& \multicolumn{1}{c|}{(f)} \\ \midrule
Baseline & 13.32 & 5.27 & 32.09 & 26.23 & 0.93 & 54.05  \\
Our ImgAny & \textbf{21.98} & \textbf{15.18} &\textbf{32.24} & \textbf{26.73}  & \textbf{37.50} & \textbf{55.61}  \\
\rowcolor{gray!10}$\Delta $  & +8.66 & +9.91 &  +0.15 & +0.50 & +36.57 & +1.56 \\
\bottomrule
\end{tabular}}
\caption{The CLIP score of the cases shown in Fig.~\ref{fig: com_2}}
\vspace{-4pt}
\label{tab: campare_3}
\end{table}

\begin{figure*}[h!]
\centering
\includegraphics[width=\textwidth]{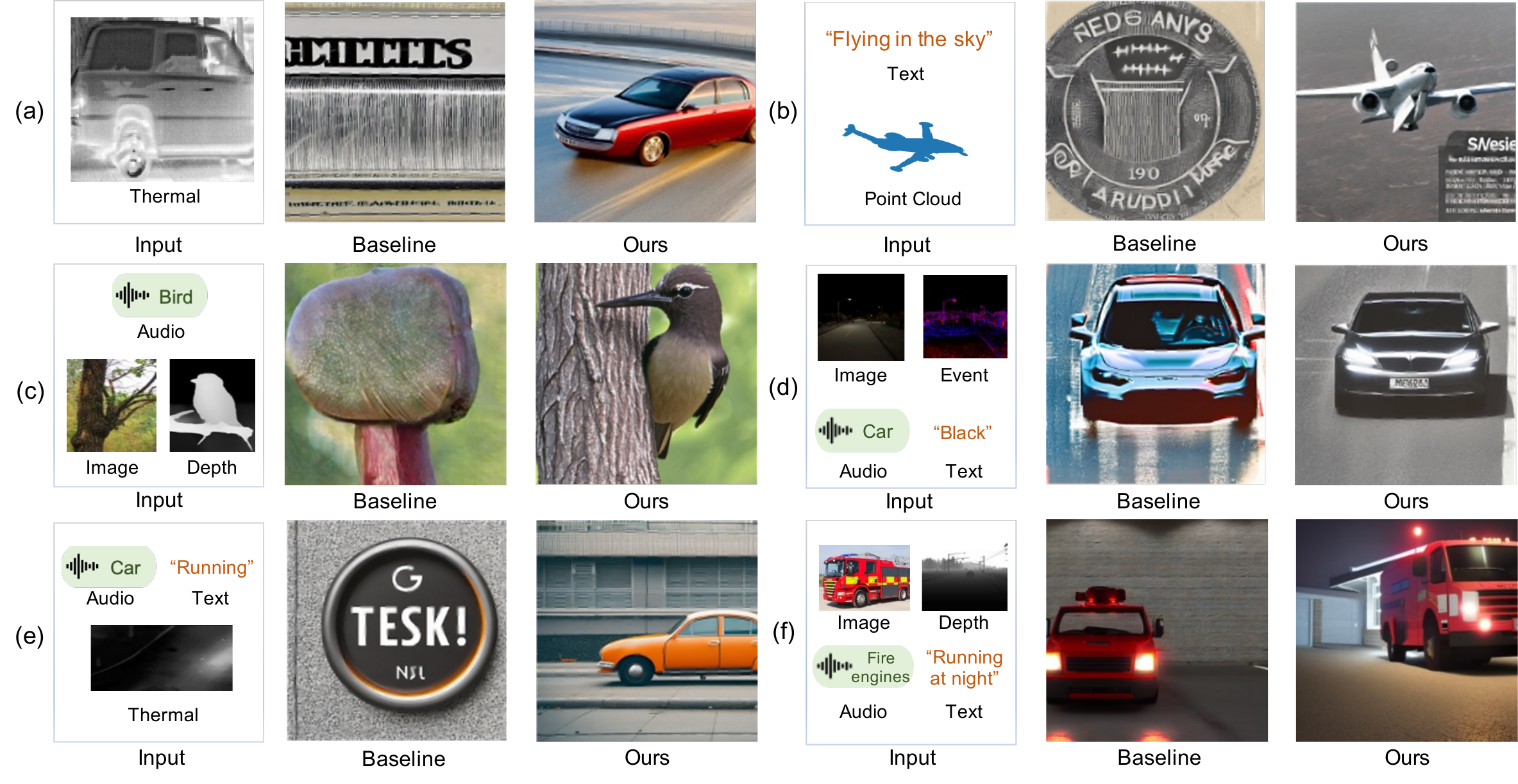}
\vspace{-8pt}
\caption{
Qualitative comparison focuses on modality combinations from seven modalities.
}
\vspace{-8pt} 
\label{fig: com_2}
\end{figure*}
\subsection{Image generation from seven Modalities}
\noindent \textbf{Qualitative Comparison.} 
As depicted in Fig.~\ref{fig: com_2}, we present a qualitative comparison of the generated results. Notably, for the modalities with considerable disparities from image modality, such as thermal, point cloud, and event modalities, our ImgAny demonstrates its efficacy in generating images from these diverse inputs. Furthermore, when processing inputs with five, six, or seven modalities, the ImgAny exhibits remarkable proficiency in extracting and retaining entity and attribute features from the multi-modal input. This capability is exemplified in the accurate depiction of complex scenarios such as "flying in the sky" and "Running at night".

\noindent \textbf{Quantitative Comparison.} 
The results in Tab.~\ref{tab: campare_3} indicate that our proposed framework, ImgAny, achieves a substantial improvement, marked by an average gain of 9.56\% in the CLIP score. This enhancement in performance underscores the superior generation stability and consistency of ImgAny, particularly in comparison to the baseline, under diverse multi-modal conditions. These findings highlight the efficacy of ImgAny in handling complex generation tasks across different combinations of modalities.

\section{Ablation Study}

\subsection{Entity Fusion Branch (EFB)}

\begin{figure*}[h!]
    \centering
    \includegraphics[width=\textwidth]{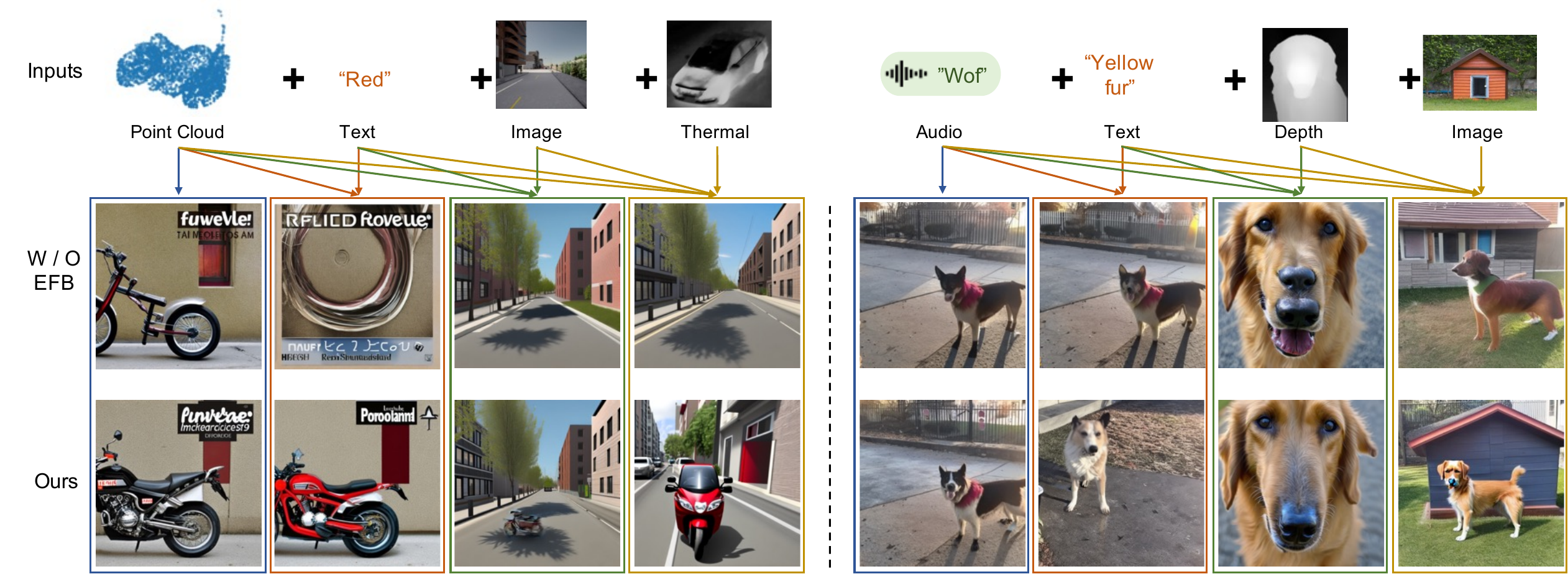}
    \caption{The qualitative comparison. We show the generation results of ImgAny with or without EFB.}
    \vspace{-4pt}
    \label{fig: ab_efb_1}
\end{figure*}

To validate the effectiveness of the entity fusion branch, we conduct ablation studies on audio-to-image and text-to-image generation. 
As Tab.~\ref{tab: ab_efb_1} indicates, the quantitative performance of ImgAny significantly decreases by 3.28\% on the CLIP score and 13.72\% on the FID score, on average, in the absence of entity fusion branch. Additionally, we present the qualitative results in Fig.~\ref{fig: ab_efb_1} to compare the quality of generated images with and without EFB. As the number of input modalities increases, more entities are lost in the generated images, which highlights the effectiveness of EFB in extracting entity features.
Tab.~\ref{tab: ab_efb_2} further illustrates the pronounced efficacy of our proposed entity fusion branch. 
This demonstrates its capability to effectively manage complex combinations of diverse modalities.

\begin{table}[t!]
\renewcommand{\tabcolsep}{12pt}
\resizebox{\linewidth}{!}{
\begin{tabular}{lcccc}
\toprule
\multicolumn{1}{c|}{Model} & \multicolumn{2}{c|}{Audio-to-Image} & \multicolumn{2}{c|}{Text-to-Image} \\ \cmidrule{2-5}
& \multicolumn{1}{|c|}{CLIP} & \multicolumn{1}{c|}{FID} & \multicolumn{1}{c|}{CLIP} & \multicolumn{1}{c|}{FID}\\ \midrule
w/o EFB & 19.97 & 248.21 &35.20  & 177.67  \\
Our ImgAny & \textbf{23.99} & \textbf{241.18 }& \textbf{37.74} & \textbf{157.26}  \\
\rowcolor{gray!10}$\Delta $  & +4.02 & +7.03 & +2.54 & +20.41   \\
\bottomrule
\end{tabular}}
\caption{Quantitative results of ImgAny with or without EFB.}
\label{tab: ab_efb_1}
\end{table}

\begin{table}[t!]
\renewcommand{\tabcolsep}{13pt}
\resizebox{\linewidth}{!}{
\begin{tabular}{lcccc}
\toprule
\multicolumn{1}{c|}{Modalities} & \multicolumn{1}{c|}{1 M} & \multicolumn{1}{c|}{2 M} & \multicolumn{1}{c|}{3 M} & \multicolumn{1}{c|}{4 M}\\ \midrule
w/o EFB & 12.91 & 18.61 & 42.68 & 23.75\\
Our ImgAny & 16.25 & 23.31 & 48.19 & 32.43 \\
\rowcolor{gray!10}$\Delta $  & 3.34 & 4.70 & 5.51 & 8.68 \\
\bottomrule
\end{tabular}}
\caption{
Performance relative to input modalities.}
\label{tab: ab_efb_2}
\vspace{-10pt}
\end{table}

\subsection{Attribute Fusion Branch (AFB)}
\begin{figure}[t!]
\centering
\includegraphics[width=\linewidth]{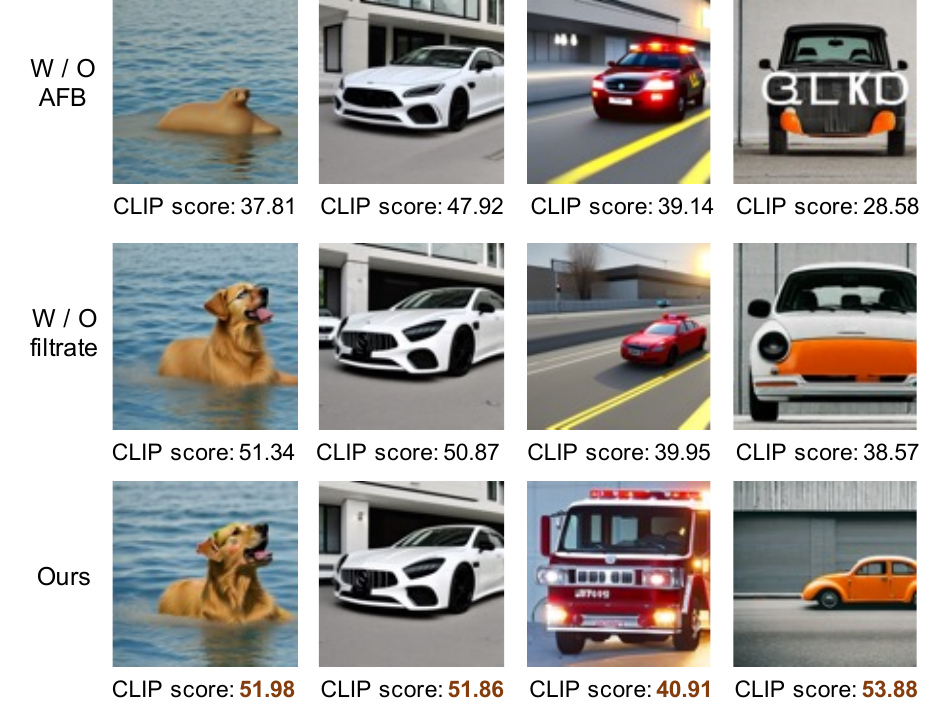}
\caption{Qualitative comparison of ImgAny with or without AFB.}
\vspace{-8pt}
\label{fig: ab_afb_1}
\end{figure}



\begin{figure}[t!]
    \centering
    \includegraphics[width=\linewidth]{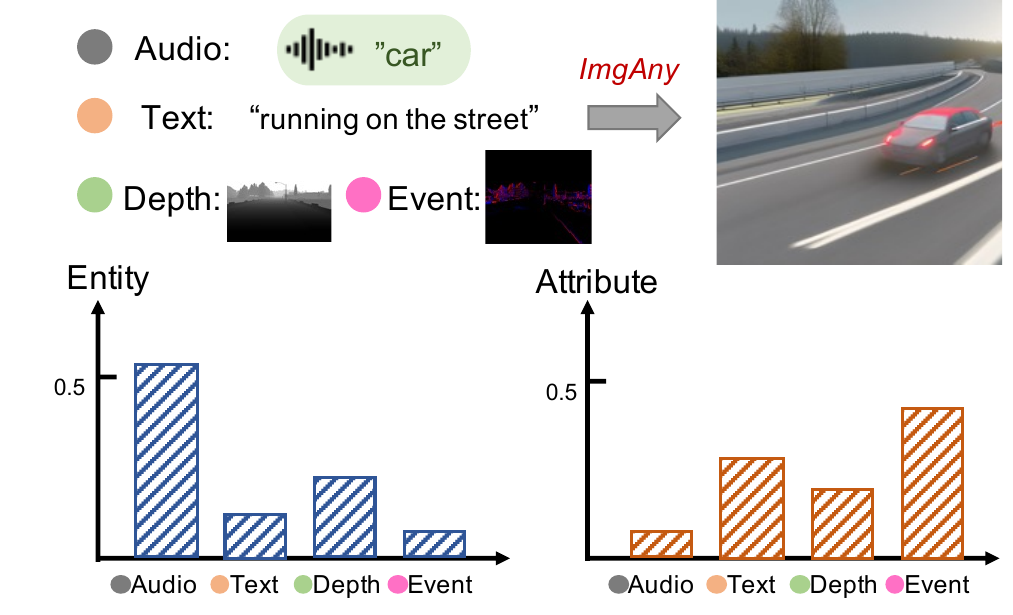}
    \vspace{-8pt}
    \caption{
    We illustrate the entity-based and attribute-based fusion weights, systematically arranging them within a two-dimensional coordinate system.
    }
    \vspace{-8pt}
    \label{fig: ab_afb_2}
\end{figure}

As shown in Fig.\ref{fig: ab_afb_1}, we present qualitative results of ImgAny with and without the attribute fusion branch. The images demonstrate that the AFB effectively extracts and retains attribute features, evident in details such as "the dog's appearance," "the car's badge (Benz)," "the state of the fire engines," and "the car's color (yellow)." Quantitative comparison results (CLIP score) in Fig.~\ref{fig: ab_afb_1} show significant improvements in image generation performance with our proposed AFB.
Furthermore, we examine the efficacy of filtrate processing within the attribute fusion branch. As demonstrated in Fig.\ref{fig: ab_afb_1}, ImgAny, when equipped with filtrate processing, attains a maximum gain of 15.31\% in generation performance on the CLIP score.
Lastly, we illustrate the entity-based and attribute-based fusion weights for various input modalities in Fig.~\ref{fig: ab_afb_2}. For case (a), within the two histograms, it is observed that the text modality predominantly contributes information to the "yellow" attribute, whereas the image and audio modalities primarily support the entity features of "dog" and "house." In case (b), the audio, depth, and event modalities contribute significantly to the entity features of "car" and "street". In contrast, the text modality primarily enhances the "running" stage in the synthesized image. These cases reveal that the entities and attributes in the generated images align with human cognitive patterns, thereby showcasing the human-like reasoning capabilities of our ImgAny.



\section{Human Evaluation}
To assess the reasoning-coherent characteristics of the generated images, human evaluation was undertaken. Data acquisition primarily revolved around participants' subjective assessments of the reasoning-coherent ability and generation quality.

\noindent \textbf{Participants.}
The study involved 27 participants, with 66.7\% aged 18-24 and 33.3\% aged 25-34. The gender distribution was 59.3\% male and 40.7\% female. Additionally, 51.9\% of the participants had previous experience with AIGC models.

\noindent \textbf{Task and Measurement.} Participants rated images using a 7-point Likert scale to evaluate coherence and depth of reasoning in relation to the inputs. Images were presented in a random order.

\begin{figure}[t!]
    \centering
    \includegraphics[width=\linewidth]{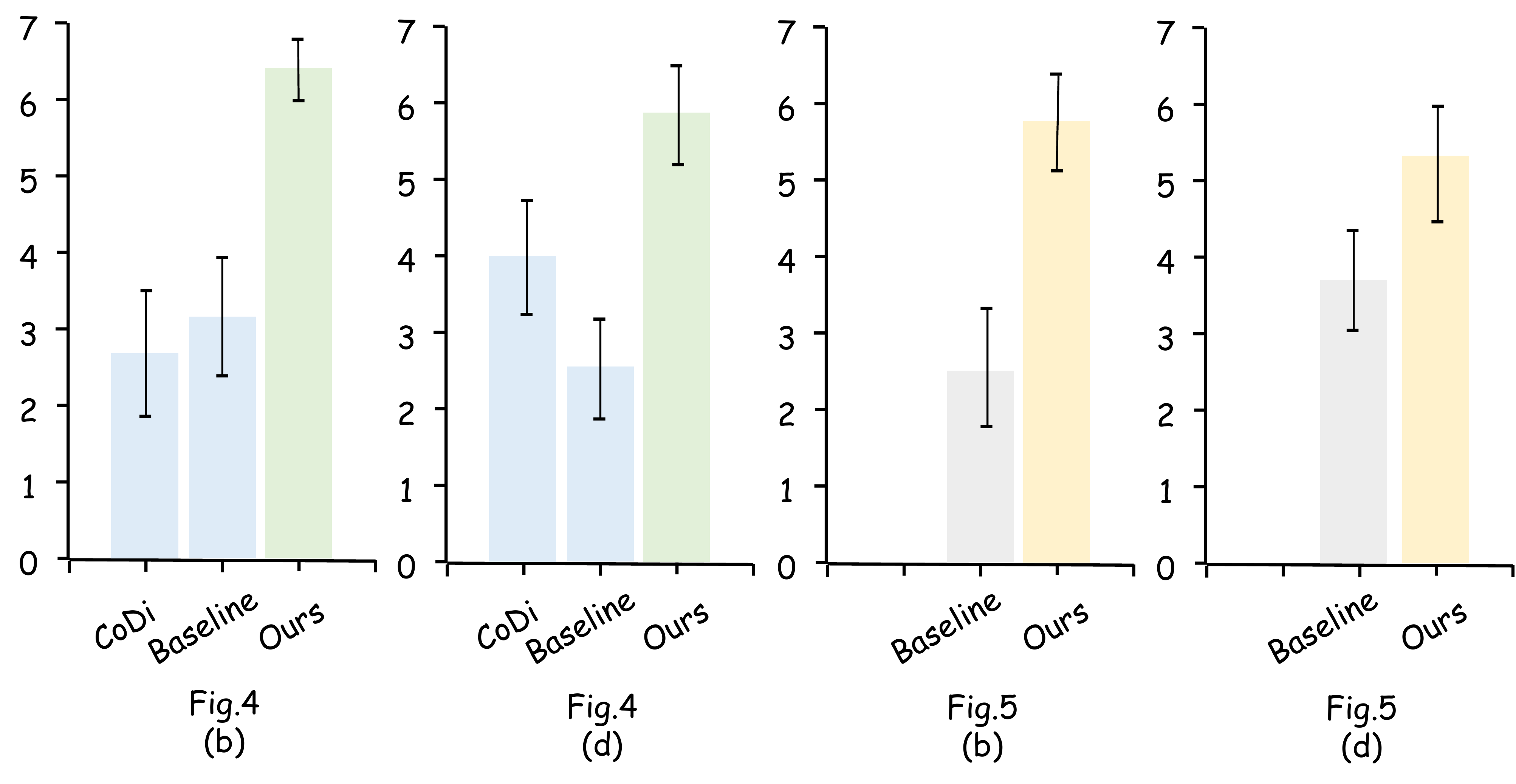}
    \caption{
    The participants' responses were assessed on a 7-point Likert scale.
    }
    \vspace{-10pt}
    \label{fig: human}
\end{figure}

\noindent \textbf{Results.} As shown in Fig.~\ref{fig: human}, all participants consistently rated the reasoning-coherent ability of ImgAny-generated images with a mean score exceeding 5, indicating significant performance gains compared to CoDi and the baseline method. ImgAny exhibited a relatively small variance across samples, highlighting its proficiency in emulating human-level reasoning and creativity.

\section{Conclusion}
In this paper, we proposed ImgAny, a training-free image generation method, characterized by its adaptability to arbitrary combinations of diverse modalities. 
ImgAny exhibits robust generation quality across any combination of modalities and demonstrates capabilities of human-level reasoning and creativity, underpinned by the integration of two fusion branches: the Entity Fusion Branch and the Attribute Fusion Branch. To assess the framework's effectiveness, we conducted comprehensive evaluations, encompassing qualitative, quantitative, and user studies. The results indicate ImgAny exhibits exceptional capability for visual creation.

\newpage
\clearpage
{
    \small
    \bibliographystyle{ieeenat_fullname}
    \bibliography{main}
}

\begin{figure*}[t!]
    \centering
    \includegraphics[width=\textwidth]{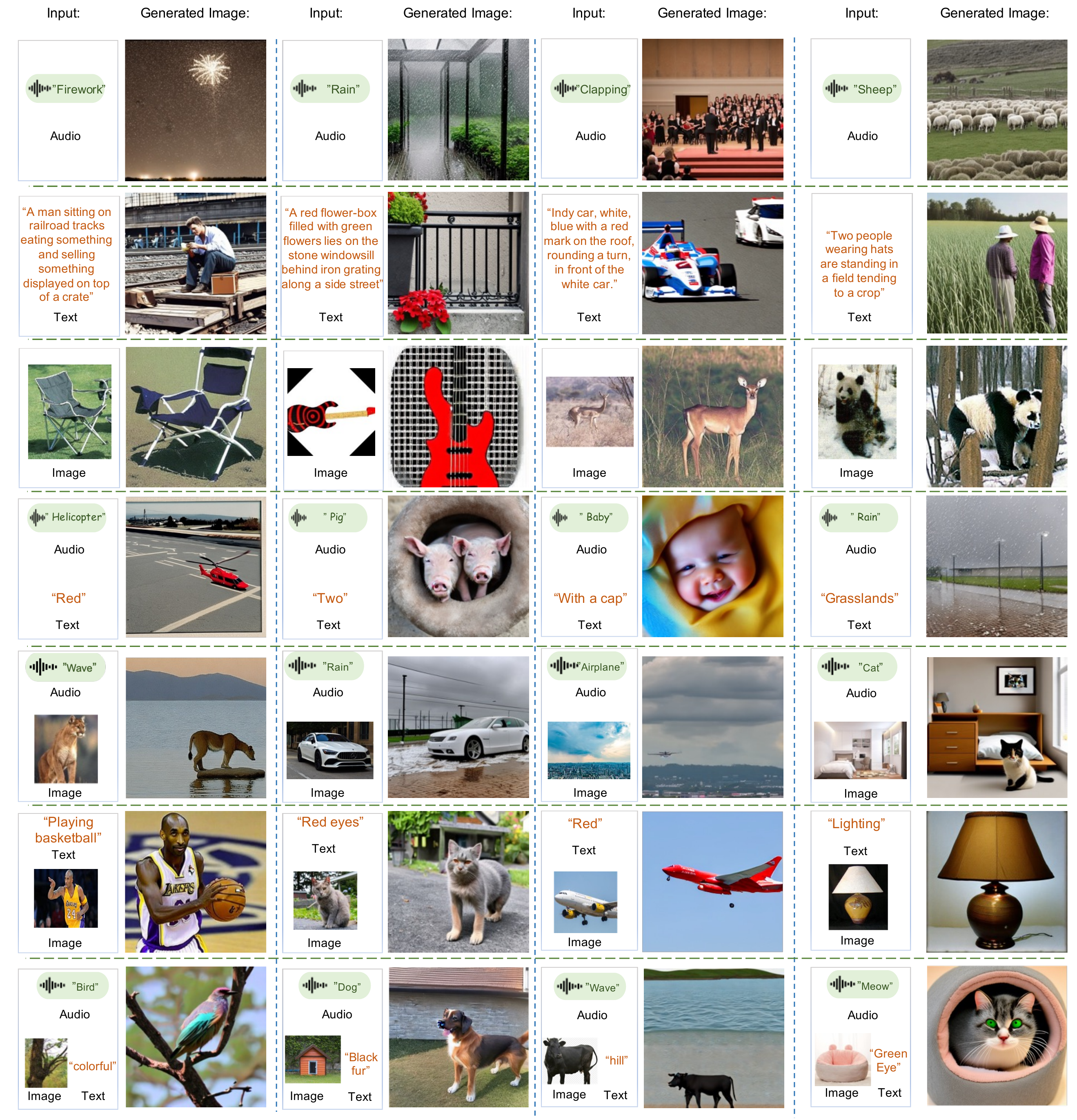}
    \caption{
    The visualization results of generated images from any combination of text, audio, and image modalities.
    }
    \label{fig: add_fig_1}
\end{figure*}

\begin{figure*}[t!]
    \centering
    \includegraphics[width=\textwidth]{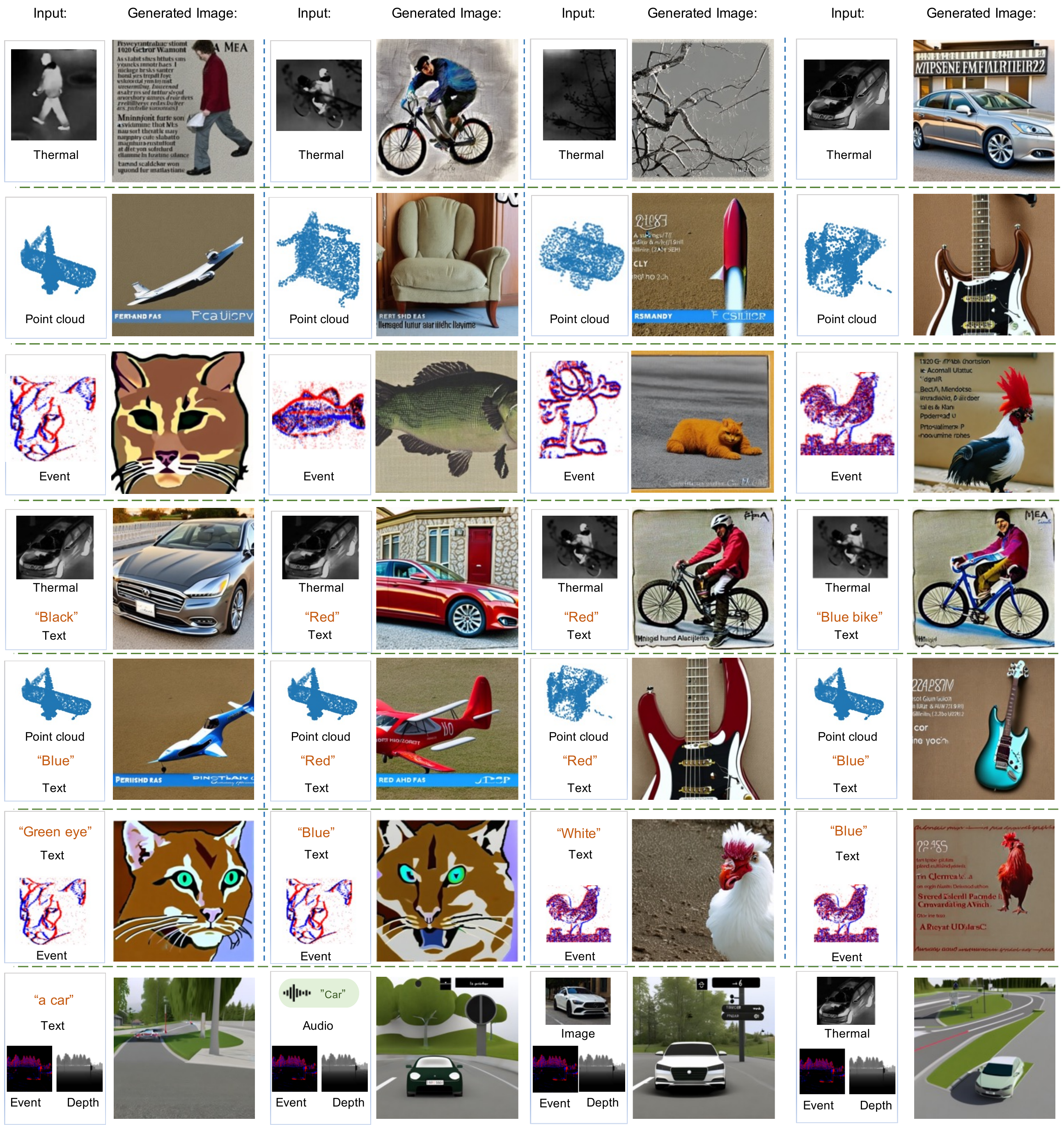}
    \caption{
    The visualization results of generated images from any combination of text, thermal, point cloud, and event modalities.
    }
    \label{fig: add_fig_2}
\end{figure*}

\end{document}